\pdfoutput=1
\documentclass[letterpaper, 10 pt, conference]{ieeeconf}
\IEEEoverridecommandlockouts
\usepackage{times}

\usepackage{multicol}
\usepackage{multirow}
\usepackage[bookmarks=true]{hyperref}
\usepackage{array}

\usepackage[font={footnotesize}]{caption}

\usepackage{amsmath}
\usepackage{amsfonts}
\usepackage{amsthm}
\usepackage{amssymb}
\usepackage{mathrsfs}  
\usepackage{bm}
\usepackage{algorithm}
\usepackage{algpseudocode}
\usepackage{dsfont} 
\usepackage{accents} 
\usepackage{tikz-cd} 
\usepackage{adjustbox} 
\usepackage{mathtools} 

\let\labelindent\relax
\usepackage{enumitem}

\usepackage{listings}
\usepackage{xcolor}
\usepackage{caption}
\usepackage{float}

\definecolor{codebg}{rgb}{0.95,0.95,0.95}
\definecolor{keywordcolor}{rgb}{0.0,0.0,0.6}   
\definecolor{typecolor}{rgb}{0.55,0.0,0.55}    
\definecolor{stringcolor}{rgb}{0.6,0.1,0.1}    
\definecolor{commentcolor}{rgb}{0.0,0.5,0.0}   
\definecolor{funcnamecolor}{rgb}{0.0,0.3,0.7}  

\usepackage{siunitx}

\usepackage{graphicx}

\usepackage{balance}

\newcommand{\comment}[1]{} 
 
\newcommand{\norm}[1]{\left\lVert#1\right\rVert} 
\newcommand\mat[1]{\begin{bmatrix}#1\end{bmatrix}} 
\newcommand\eqs[1]{\begin{equation}\begin{split}#1\end{split}\end{equation}} 
\newcommand\eqsnn[1]{\begin{equation*}\begin{split}#1\end{split}\end{equation*}} 
\newcommand\eqa[1]{\begin{align}#1\end{align}} 

\newcommand\parens[1]{\left(#1\right)} 

\newcommand\R{\mathbb{R}} 
\newcommand\koop{\bm{\mathcal{K}}}

\newcommand\state{\bm{x}} 
\newcommand\latent{\bm{z}} 
\newcommand\statechunk{\bm{\xi}} 
\newcommand\latentdelay{\bm{\zeta}} 
\newcommand\latentdyn{\bm{K}} 
\newcommand\decoder{\bm{h}} 
\newcommand\statedyn{\bm{f}} 
\newcommand\encoder{\bm{g}} 
\newcommand\kfmean{\bm{\mu}} 
\newcommand\kfcov{\bm{\Sigma}} 
\newcommand\kfmeanol{\bar{\bm{\mu}}} 
\newcommand\kfcovol{\bar{\bm{\Sigma}}} 
\newcommand\kalmangain{\bm{G}} 
\newcommand\procnoise{\bm{Q}} 
\newcommand\measnoise{\bm{R}} 
\newcommand\decoderjac{\bm{C}} 
\newcommand\numdelays{n_z} 
\newcommand\chunksize{n_x} 
\newcommand\latentsize{d_z} 
\newcommand\koopblock{\Gamma} 
\newcommand\loss{\mathcal{L}} 
\newcommand\embedspace{\mathcal{Z}} 
\newcommand\statespace{\mathcal{X}} 
\newcommand\prob{p}
\newcommand\embeddingfunctionspace{\mathscr{G}}

\usepackage{pifont}

\usepackage{lipsum}
\usepackage{booktabs}

\newcommand{\algcommentfont}{\ttfamily\color{green!50!black}}
\makeatletter

\algrenewcommand{\algorithmiccomment}[1]{\hfill{\algcommentfont \#~#1}}
\makeatother
\AddToHook{env/algorithmic/begin}{\small}

\begin{document}

\title{\LARGE \bf KALIKO: \underline{Kal}man-\underline{I}mplicit \underline{K}oopman \underline{O}perator Learning \\
For Prediction of Nonlinear Dynamical Systems}
\author{
    Albert H. Li$^*$\thanks{* Equal contribution. A. H. Li is supported by Dow Award 227027AW. Authors are all with Caltech.}, Ivan Dario Jimenez Rodriguez$^*$, Joel W. Burdick, Yisong Yue, Aaron D. Ames
}

\maketitle

\begin{abstract}

Long-horizon dynamical prediction is fundamental in robotics and control, underpinning canonical methods like model predictive control. Yet, many systems and disturbance phenomena are difficult to model due to effects like nonlinearity, chaos, and high-dimensionality.
Koopman theory addresses this by modeling the linear evolution of embeddings of the state under an infinite-dimensional linear operator that can be approximated with a suitable finite basis of embedding functions, effectively trading model nonlinearity for representational complexity. However, explicitly computing a good choice of basis is nontrivial, and poor choices may cause inaccurate forecasts or overfitting. To address this, we present Kalman-Implicit Koopman Operator (KALIKO) Learning, a method that leverages the Kalman filter to \textit{implicitly} learn embeddings corresponding to latent dynamics without requiring an explicit encoder. KALIKO produces interpretable representations consistent with both theory and prior works, yielding high-quality reconstructions and inducing a globally linear latent dynamics. Evaluated on wave data generated by a high-dimensional PDE, KALIKO surpasses several baselines in open-loop prediction and in a demanding closed-loop simulated control task: stabilizing an underactuated manipulator's payload by predicting and compensating for strong wave disturbances.

\end{abstract}

\IEEEpeerreviewmaketitle


\section{Introduction}\label{sec:intro}

Dynamical modeling and prediction are fundamental problems in robotics and control, forming the bedrock for techniques like model predictive control, state estimation, and more. However, obtaining accurate analytical models in unstructured settings can be challenging, especially for nonlinear systems that may be subject to high-dimensional and/or chaotic spatiotemporal phenomena like turbulence or surface waves. This motivates a large body of work on fitting predictive models from data \cite{brunton2020machine,brunton2022_koopmanreview,otto2019_linearlyrecurrentautoencodernetworkslearning, shi2025_koopmanoperatorsrobotlearning}.

\begin{figure}[t]
    \vspace{0.2cm}
    \begin{center}
        \includegraphics[width=\linewidth]{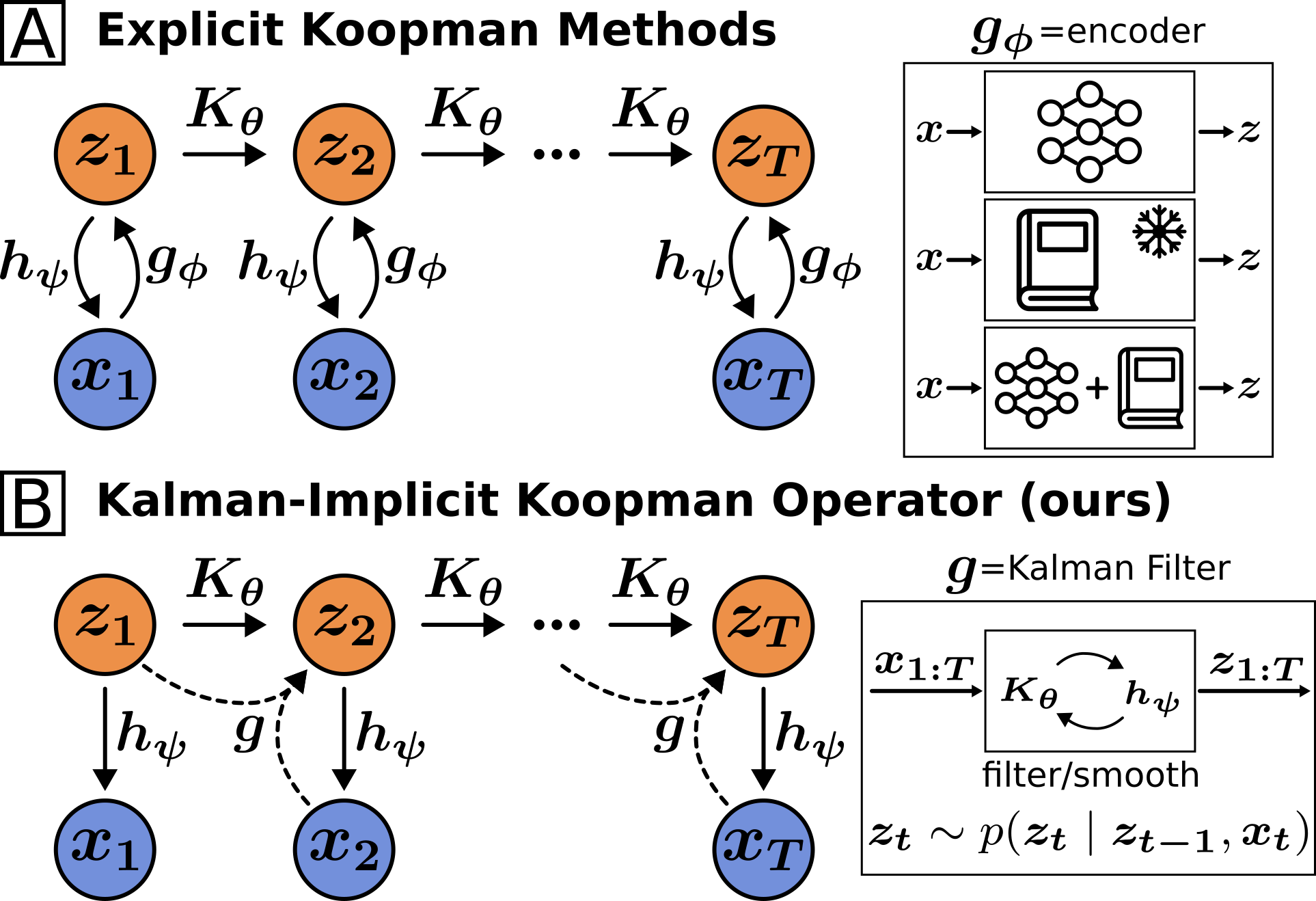}
    \end{center}
    \caption{
        \textbf{(A)} Most methods for learning the linear Koopman dynamics $\latentdyn_\theta$ are \textit{explicit}, parameterizing an encoder $\encoder_\phi$ that maps data $\state$ to high-dimensional Koopman embeddings $\latent$. Examples include neural networks, fixed dictionaries of basis functions, or learnable dictionaries. \textbf{(B)} KALIKO (ours) instead \textit{implicitly} learns Koopman embeddings by letting $\encoder$ be a Kalman filter and smoother, which is composed of only two models: the latent dynamics $\latentdyn_\theta$ and decoder $\decoder_\psi$. We show that these implicitly-recovered representations are highly interpretable and yield strong open-loop prediction and closed-loop control performance.
    }
    \label{fig:hero}
    \vspace{-0.5cm}
\end{figure}

One appealing approach is Koopman operator theory, which models the \textit{linear} evolution of state embeddings\footnote{We use the term ``embedding'' rather than ``observable'' to avoid confusion with ``measurements'' and ``observations'' in Kalman filtering.} under the infinite-dimensional \textit{Koopman operator} $\koop$. In this lifted space, nonlinear dynamics become linear, effectively trading nonlinearity for representational complexity \cite{brunton2022_koopmanreview}. In practice, one builds finite approximations of the Koopman operator by \textit{explicitly} parameterizing embedding functions via black-box neural networks \cite{lusch2018_deepkoopman, otto2019_linearlyrecurrentautoencodernetworkslearning}, fixed ``dictionaries'' of basis functions (e.g., polynomials) \cite{brunton2016_sindy}, or hybrid approaches that learn these dictionaries \cite{jin2024_invertibledictionary} (see Fig.~\ref{fig:hero}A). However, selecting embeddings that capture an approximately invariant subspace is difficult; poor choices produce spurious modes or overfitted models and often necessitate multi-objective losses with carefully tuned weights \cite{lusch2018_deepkoopman, yeung2019_deepdmd, otto2019_linearlyrecurrentautoencodernetworkslearning}.

This work presents an alternative approach: \textit{\underline{\textbf{Kal}}man-\underline{\textbf{I}}mplicit \underline{\textbf{K}}oopman \underline{\textbf{O}}perator} (KALIKO) Learning. Unlike these explicit approaches, KALIKO reframes the problem of encoding data $\state$ into embeddings $\latent$ as a Bayesian state estimation problem that infers latent states $\latent_{1:T}$ from data trajectories $\state_{1:T}$. Thus, KALIKO \textit{implicitly} represents an encoder via an (extended) Kalman filter, allowing it to circumvent the challenge of parameterizing an embedding function (see Fig. \ref{fig:hero}B). We show that despite its implicit structure, KALIKO learns highly interpretable dynamics consistent with Koopman theory and prior works (e.g., eigenfunctions for orbits, limit-cycles, and equilibria), achieves superior prediction performance over several baselines, and translates its capabilities to a challenging closed-loop stabilization task.


\subsection{Related Work}\label{sec:related_work}

\textbf{Computing Koopman Embeddings.} A standard data-driven method is \textit{Dynamic Mode Decomposition} (DMD) \cite{rowley2009_dmd, brunton2022_koopmanreview}, which fits a least-squares linear map between state snapshots, projecting $\koop$ onto the span of embedding coordinates of those snapshots. As DMD operates on raw measurements, it cannot in general realize a nonlinear change of coordinates; \textit{Extended DMD} (EDMD) addresses this by regressing in a lifted space built from a \emph{dictionary} of embedding functions, like polynomials, which must be carefully constructed \cite{williams2015data, brunton2022_koopmanreview}. Time-delay coordinates provide another practical lifting technique: Delay-DMD constructs Hankel embeddings from past samples to capture approximately Koopman-invariant structure \cite{brunton2017_havok}. Yet another method is \textit{Kernel DMD} (KDMD), which implements EDMD \textit{implicitly} by only learning inner products between the embedded data, avoiding explicit enumeration of the dictionary \cite{otto2019_linearlyrecurrentautoencodernetworkslearning, meng2024_koopmaninvertiblenn}. Lastly, data-driven methods attempt to learn these embeddings, either by learning dictionary elements or by training a deep autoencoder alongside a latent dynamics model \cite{lusch2018_deepkoopman, takeishi2018_learningkoopmaninvariantsubspaces, yeung2019_deepdmd, otto2019_linearlyrecurrentautoencodernetworkslearning, jin2024_invertibledictionary}.

These diverse approaches share canonical failure modes. Even for linear systems, a single spurious basis function can induce severe overfitting \cite{otto2019_linearlyrecurrentautoencodernetworkslearning}. In the data-driven setting, practitioners counter this with heavy regularization \cite{otto2019_linearlyrecurrentautoencodernetworkslearning, yeung2019_deepdmd}, complex multi-term losses \cite{lusch2018_deepkoopman, miller2022_eigenloss}, and/or highly-sensitive weight initializations and hyperparameters \cite{miller2022_eigenloss}. Conversely, convergence of learned dictionaries is often unclear \cite{yeung2019_deepdmd} and poor choices of the above can just as easily trap solutions in bad local minima, leading to underfitting \cite{miller2022_eigenloss, nayak2025_temporallyconsistentkaes}.

\textbf{Kalman Filters and Dynamical Learning.}
Learnable Kalman filters (KFs) jointly train a latent dynamics model and decoder that maximize the likelihood of data trajectories. The decoder $p(\state_t \mid \latent_t)$ is trained together with the dynamics $p(\latent_{t} \mid \latent_{t-1})$ so that new data points $\state_t$ accurately update the posterior belief over the latent state $\latent_t$. Unlike autoencoder-based approaches that learn an encoder/decoder independently of the dynamics, KFs couple representation and prediction by \textit{interleaving} prediction and measurement updates (Sec. \ref{sec:prelims_kalman}), which encourages embeddings suitable for both long-horizon prediction and reconstruction.

Prior work has leveraged this insight to learn latent dynamics for image sequences \cite{fraccaro2017_kvae, li2021_replayovershooting} by \textit{compressing} high-dimensional observations into a low-dimensional latent state. In line with Koopman theory, KALIKO instead \textit{lifts} the dimension to find an embedding space where the latent dynamics are approximately linear. Moreover, almost all prior works learn filters for state estimation \cite{haarnoja2017_backpropkf, lee2020_multimodalsensorfusion, kloss2021_difffilters, corenflos2021_differentiablepf} or simple open-loop prediction tasks \cite{fraccaro2017_kvae, li2021_replayovershooting}. In this work, we show that KALIKO's predictions can also be used in a closed-loop control task.

\textbf{Kalman Filters and Koopman.} Many works have combined KFs with Koopman methods. Most commonly, KFs are used for latent state estimation \textit{after} learning some linear dynamics \cite{surana2016_koopmanobserver, zeng2023_koopmankfcosseratrod, huang2024_koopmanekfquadrotor}. Others have used KFs for spectral analysis \cite{liu2024_estimatekoopmanmodeseigenvalues, zeng2025_koopmanspectralanalysisnoisy}.
Closest to our work are system identification or operator learning methods, but unlike KALIKO, assume the existence of embeddings \textit{a priori} \cite{guan2025_koopmansysidsphericalrobot} or train an encoder rather than a decoder \cite{guo2021_koopmanlinearizationdatadrivenbatch}.

\textbf{Time Series Prediction.} The machine learning community has used methods like recurrent neural networks \cite{sherstinsky2020_rnns} to model sequential data and learn dynamical systems \cite{mohajerin2018_rnndynamics}, especially those that are hard to model, like continuum robot dynamics \cite{tariverdi2021_rnncontinuumrobot}, human motion \cite{zhang2020_rnnhumanmotion}, and video \cite{hafner2019_planet}. More recently, transformer-based models have been applied to many phenomena like weather, traffic patterns, electricity usage, etc. \cite{nie2023_patchtst}. It has been shown that LLM backbones like GPT2 are effective for time series prediction \cite{zhou2023_onefitsallgpt2}, prompting research on time series tokenization \cite{talukder2025_totem}.

The recent work ``Koopman-Kalman Enhanced Variational Autoencoder'' (K2VAE) at first appears similar to KALIKO, but several details distinguish them. First, K2VAE learns an explicit encoder. Second, K2VAE uses its KF to drive learning error to 0 with a dynamics $\mathbf{A}$ distinct from $\koop$, while KALIKO uses the KF directly on the learned operator. Lastly, KALIKO uses several techniques not present in K2VAE that are crucial for performance (see Sec. \ref{sec:impl} and \ref{sec:ablations}). We compare to K2VAE in Sec. \ref{sec:experiments_prediction}.

\subsection{Contributions}\label{sec:contributions}
Our chief contribution is KALIKO, a method that uses the Kalman filter to \textit{implicitly} learn a Koopman embedding space governed by globally linear latent dynamics. Second, we show that KALIKO's embeddings produce faithful reconstructions, are highly interpretable, and align with theoretical expectations. Lastly, on complex wave data generated by a high-dimensional PDE, KALIKO outperforms several baselines on both an open-loop prediction task and a challenging simulated closed-loop stabilization task where it must stabilize an underactuated manipulator's payload by predicting and compensating for wave disturbances.


\section{Mathematical Preliminaries}\label{sec:preliminaries}
Let $\state \in \statespace \subseteq \R^{n}$ be the state of a discrete-time nonlinear system which generates the data of interest,
\begin{align}
    \state_{t+1} = \statedyn(\state_t), \label{eq:dyn_data}
\end{align}
and $\latent \in \embedspace \subseteq \R^{m}$ be the state of some corresponding desired linear latent dynamics
\eqs{\label{eq:dyn_latent}
    \latent_{t+1} = \latentdyn_{\theta} \latent_t, \quad \state_t = \decoder_\psi(\latent_t),
}
where $\theta,\psi$ are learnable parameters and $\decoder: \embedspace \to \statespace$ decodes the latent state back to the data domain. We use ``encoder'' or ``embedding function'' to refer to any operation $\encoder:\statespace\rightarrow\embedspace$ mapping data $\state$ to latents $\latent$.

\begin{figure*}[t]
    \vspace{0.15cm}
    \begin{center}
        \includegraphics[width=\linewidth]{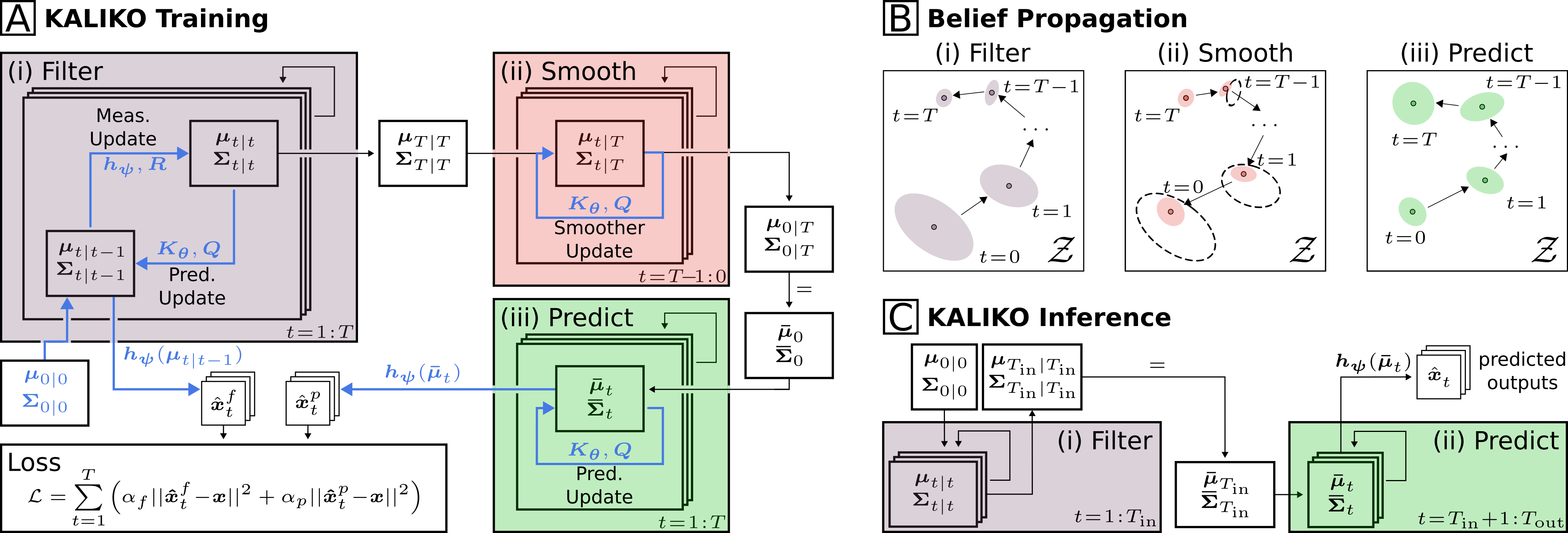}
    \end{center}
    \caption{
        \textbf{(A) KALIKO Training.} (i) Filter for $T$ steps, outputting a sequence of filtered distributions $\{(\kfmean_{t \mid t-1}, \kfcov_{t \mid t-1})\}_{t=1}^T$ used for the loss. (ii) Run a backward smoother from the final belief $(\kfmean_{T \mid T}, \kfcov_{T \mid T})$, yielding the posterior initial belief $(\kfmean_{0\mid T}, \kfcov_{0\mid T})$. (iii) Roll this belief forward for $T$ steps to get predicted beliefs $\{(\bar\mu_t,\bar\Sigma_t)\}_{t=1}^T$. Filtered and predicted means are decoded to observations and the model is trained end-to-end with a reconstruction loss. Learnable modules/parameters and their operations are highlighted in blue. \textbf{(B) Belief Propagation.} The filter's latent belief becomes more certain with more steps. The smoother reuses future evidence to calibrate the initial belief, seeding a strong prediction rollout during training. \textbf{(C) KALIKO Inference.} KALIKO filters on $T_\text{in}$ data points then rolls out its latent dynamics for $T_\text{out}$ steps. The $T_\text{out}$ predicted latents are decoded into the final predicted trajectory.
    }
    \label{fig:method}
    \vspace{-0.5cm}
\end{figure*}

\subsection{Koopman Theory}\label{sec:prelims_koopman}

We provide a brief overview of Koopman theory, deferring a detailed treatment to \cite{brunton2022_koopmanreview}. Let $\embeddingfunctionspace(\statespace)$ be a vector space of embedding functions on data $\state \in \statespace$.
The \textit{Koopman operator} $\koop:\embeddingfunctionspace(\statespace) \to \embeddingfunctionspace(\statespace)$ acts by composition,
\eqs{
(\koop \encoder)(\state)=\encoder\big(\statedyn(\state)\big),\quad \encoder\in\embeddingfunctionspace(\statespace),
}
and is linear even if $\statedyn$ is nonlinear. Since $\koop$ is generally infinite-dimensional, in practice we seek finite approximations $\latentdyn \in \R^{m \times m}$ of $\koop$ as in \eqref{eq:dyn_latent}, and we say $\encoder$ is \textit{$\koop$-invariant} if the following closure condition is satisfied:
\eqs{\label{eq:koop_closure}
    \encoder(\statedyn(\state)) = \latentdyn\encoder(\state),\quad\forall\state.
}
That is, the latents $\latent$ evolve linearly on the range of $\encoder$. When $\eqref{eq:koop_closure}$ fails, $\latentdyn$ introduces \textit{spurious modes} that accumulate multi-step prediction error due to the closure residual \cite{otto2019_linearlyrecurrentautoencodernetworkslearning, brunton2016_sindy, brunton2022_koopmanreview}. Thus, computing and analyzing invariant embeddings is central to Koopman analysis, which we revisit in Sec. \ref{sec:embeddings}.

\subsection{Kalman Filtering}\label{sec:prelims_kalman}

We now review the mechanics of Kalman filters (KFs), deferring details to \cite{särkkä2013_filtering}. KFs recursively update a posterior belief ${\prob(\latent_t \mid \latent_{t-1}, \state_t)}$ over the state given new measurements and its previous state. We assume Gaussian noise with covariance $\procnoise$ corrupts the dynamics such that
\eqs{\label{eq:transition_distribution}
    \prob(\latent_t \mid \latent_{t-1}) = \mathcal{N}(\latentdyn\latent_{t-1}, \procnoise),
}
and similarly, that Gaussian noise with covariance $\measnoise$ corrupts the decoder such that
\eqs{\label{eq:decoder}
    \prob(\state_t \mid \latent_t) = \mathcal{N}(\decoder(\latent_t), \measnoise).
}

Starting from a prior belief over $\latent_0 \sim \mathcal{N}(\kfmean_{0\mid0}, \kfcov_{0\mid0})$, the KF iteratively updates its belief by interleaving two updates. 
First, the \textit{prediction update} propagates the belief at time $t-1$ through the dynamics:
\eqa{
    \kfmean_{t \mid t-1} &= \latentdyn \kfmean_{t-1 \mid t-1}, \label{eq:prediction_update_1} \\
    \kfcov_{t \mid t-1} &= \latentdyn \kfcov_{t-1 \mid t-1} \latentdyn^\top + \procnoise \label{eq:prediction_update_2}.
}
Second, the \textit{measurement update} adjusts the posterior by accounting for the likelihood of observing $\state_t$:
\eqa{
    \kalmangain_t &= \kfcov_{t \mid t-1}\decoderjac_t^\top\parens{\decoderjac_t\kfcov_{t \mid t-1}\decoderjac_t^\top + \measnoise}^{-1}, \label{eq:meas_update_1} \\
    \kfmean_{t \mid t} &= \kfmean_{t \mid t-1} + \kalmangain_t\parens{\state_t - \decoder(\kfmean_{t \mid t-1})}, \label{eq:meas_update_2} \\
    \kfcov_{t \mid t} &= (I - \kalmangain_t\decoderjac_t)\kfcov_{t \mid t-1}, \label{eq:meas_update_3}
}
where $\decoderjac_t$ is the decoder's Jacobian, $\decoderjac_t := \frac{\partial}{\partial \latent}\decoder(\kfmean_{t \mid t-1})$.

The Kalman filter causally updates its belief distribution by observing only past data. 
However, after observing $T$ data points, it is often useful to update \textit{past} beliefs conditioned on future observations using \textit{Kalman smoothing}. 
The Rauch-Tung-Striebel smoother caches values from the filtering pass to recursively compute smoothed beliefs from time $T$ to $0$:
\eqa{
    \kalmangain_t^s &= \kfcov_{t \mid t}\latentdyn^\top\parens{\kfcov_{t + 1 \mid t}^\top}^{-1}, \label{eq:smoothing_1} \\
    \kfmean_{t \mid T} &= \kfmean_{t \mid t} + \kalmangain_t^s(\kfmean_{t+1 \mid T} - \kfmean_{t+1 \mid t}), \label{eq:smoothing_2} \\
    \kfcov_{t \mid T} &= \kfcov_{t \mid t} + \kalmangain_t^s(\kfcov_{t+1\mid T} - \kfcov_{t+1 \mid t})\parens{\kalmangain_t^s}^\top. \label{eq:smoothing_3}
}

In summary, filtering and smoothing require the dynamics $\latentdyn$, decoder $\decoder$, prior belief $(\kfmean_{0 \mid 0}, \kfcov_{0 \mid 0})$, dynamics and measurement covariances $\procnoise$ and $\measnoise$, and a sequence of data points $\state_{1:T}$. In return, they yield filtered and smoothed belief distributions over the corresponding latent states $\latent_{1:T}$, so we can view the Kalman filter/smoother as an implicit encoder over trajectories, $\encoder: \state_{1:T} \mapsto \latent_{1:T}$.


\section{The Kalman-Implicit Koopman Operator}\label{sec:kaliko}

Our method has two goals: (i) producing high-quality predictions under latent linear dynamics $\latentdyn$, and (ii) accurately decoding them with $\state=\decoder_\psi(\latent)$. Observe that inaccurate dynamics $\latentdyn$ degrade the prediction update \eqref{eq:prediction_update_1}–\eqref{eq:prediction_update_2}, and an inaccurate decoder $\decoder_\psi$ degrades the measurement update \eqref{eq:meas_update_1}–\eqref{eq:meas_update_3}. Thus, a performant KF requires both accurate $\latentdyn$ and $\decoder_\psi$. This motivates our method, Kalman-implicit Koopman operator (KALIKO) learning, which trains a KF end-to-end with a single reconstruction objective to jointly achieve both goals. In contrast, autoencoding Koopman methods typically separate prediction and reconstruction \cite{lusch2018_deepkoopman, otto2019_linearlyrecurrentautoencodernetworkslearning, meng2024_koopmaninvertiblenn}, which may overemphasize one at the expense of the other.

\subsection{Training}\label{sec:kaliko_training}

KALIKO is trained with the replay overshooting method \cite{li2021_replayovershooting}, which consists of three stages (see Fig. \ref{fig:method}A). Given a data trajectory $\state_{1:T}$, the \textit{filtering stage} interleaves the prediction and measurement updates in \eqref{eq:prediction_update_1}-\eqref{eq:prediction_update_2} and \eqref{eq:meas_update_1}-\eqref{eq:meas_update_3} respectively to recover the sequences of belief distributions $\{\parens{\kfmean_{t \mid t-1}, \kfcov_{t \mid t-1}}\}_{t=1}^T$ and $\{\parens{\kfmean_{t \mid t}, \kfcov_{t \mid t}}\}_{t=1}^T$. The \textit{smoothing stage} then applies \eqref{eq:smoothing_1}-\eqref{eq:smoothing_3} to these beliefs in reverse to recover an initial belief at $t=0$ conditioned on all measurements, $(\kfmean_{0 \mid T}, \kfcov_{0 \mid T})$. Lastly, the \textit{prediction stage} rolls out $T$ prediction updates \eqref{eq:prediction_update_1}-\eqref{eq:prediction_update_2} from the smoothed initial condition \textit{without measurement updates} to recover a sequence of predicted beliefs $\{\parens{\kfmeanol_t, \kfcovol_t}\}_{t=1}^T$.

KALIKO learns $\latentdyn_\theta$ and $\decoder_\psi$ along with the prior ($\kfmean_{0\mid 0}$, $\kfcov_{0\mid0}$) and covariances ($\procnoise$, $\measnoise$) end-to-end via the loss
\eqs{\label{eq:loss}
    \resizebox{0.88\linewidth}{!}{
        \ensuremath{
            \loss = \sum_{t=1}^T \alpha_f \underbrace{\norm{\decoder_\psi\parens{\kfmean_{t \mid t-1}} - \state_t}_2^2}_{\text{filtering}} + \alpha_p \underbrace{\norm{\decoder_\psi\parens{\kfmeanol_t} - \state_t}_2^2}_{\text{prediction}},
        }
    }
}
where the first term corresponds to the filtered\footnote{We use the pre-measurement update mean $\kfmean_{t\mid t-1}$, as in \cite{li2021_replayovershooting}.} reconstructions of the data and the second to the predicted reconstructions. The coefficients $\alpha_f,\alpha_p\in\R_{>0}$ control the relative weights for filtering and prediction respectively, which we set to $1$ in all experiments.

The \textit{filtering loss} is analogous to an autoencoding loss, as it fuses measurement consistency (``decoding'') via \eqref{eq:meas_update_1}-\eqref{eq:meas_update_3} with latent inference via \eqref{eq:prediction_update_1}-\eqref{eq:prediction_update_2} (``encoding'') \cite{li2021_replayovershooting}.
The \textit{prediction loss} ensures accurate open-loop long-horizon predictions, which mirrors inference-time execution. We remark that \cite{li2021_replayovershooting} maximizes the log-likelihood of the data while we found that a simple MSE loss was more computationally-efficient with no performance degradation.

\begin{figure}[t]
    \vspace{0.15cm}
    \begin{center}
        \includegraphics[width=\linewidth]{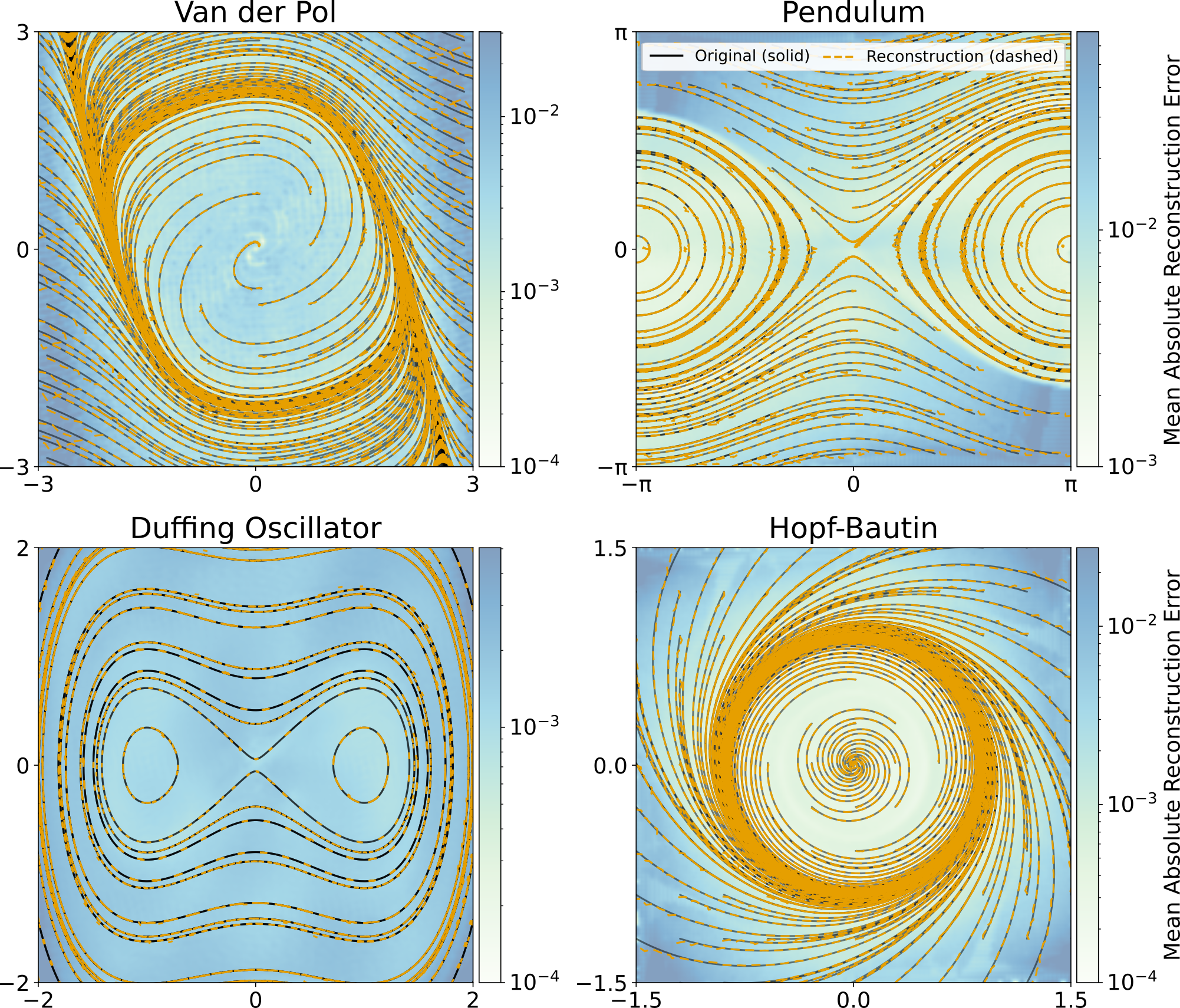}
    \end{center}
    \caption{\textbf{KALIKO Reconstructions.} KALIKO closely reconstructs the vector field of nonlinear systems without an encoder using globally linear latent dynamics. Shown are some ground truth trajectories in black (solid) and their reconstructions in orange (dashed). A denser heatmap of the reconstruction error is shown in blue (darker denotes more error).}
    \label{fig:toy_recon}
    \vspace{-0.5cm}
\end{figure}

\subsection{Inference}

At inference time, KALIKO uses $T_\text{in}$ data points to predict the next $T_\text{out}$ steps. First, the filter consumes $\state_{1:T_\text{in}}$ to obtain the posterior belief $(\kfmean_{T_\text{in} \mid T_\text{in}}, \kfcov_{T_\text{in} \mid T_\text{in}})$. $T_\text{out}$ prediction steps are then executed from this filtered belief, yielding a sequence of predicted latent beliefs $\{\parens{\kfmeanol_t, \kfcovol_t}\}_{t=T_\text{in}+1}^{T_\text{in}+T_\text{out}}$. The predicted means are each decoded to obtain a predicted trajectory: $\hat{\state}_t = \decoder_\psi(\kfmeanol_t)$. See Fig. \ref{fig:method}C for a visualization.

\subsection{Additional Implementation Details}\label{sec:impl}

\textbf{Delay Embedding.} Inspired by the use of delay embeddings in Koopman methods \cite{brunton2017_havok}, KALIKO parameterizes $\latentdyn_\theta$ as the dynamics on a delay-embedded latent state, yielding a sparse block-companion matrix form
\eqsnn{
    \underbrace{\mat{
        \latentdelay_{t + 1} \\ \vdots \\ \latentdelay_{t + \numdelays - 1} \\ \latentdelay_{t + \numdelays}
    }}_{\latent_{t+1}}%
    = \underbrace{\mat{
        0 & I & 0 & \dots & 0 \\
        \vdots & \vdots & \ddots & \ddots & \vdots \\
        0 & 0 & \dots & 0 & I \\
        \koopblock_1 & \koopblock_2 & \dots & \dots & \koopblock_{\numdelays}
    }}_{\latentdyn}%
    \underbrace{\mat{
        \latentdelay_t \\ \vdots \\ \latentdelay_{t + \numdelays - 2} \\ \latentdelay_{t + \numdelays - 1}
    }}_{\latent_t},
}
where the only learnable parameters are the blocks $\koopblock_1,\dots,\koopblock_{\numdelays}$, and each $\latentdelay_t$ is a sub-latent state of dimension $\latentsize$, which is stacked into the full delay-embedded latent state $\latent_t = \left(\latentdelay_t, \ldots, \latentdelay_{t+\numdelays-1}\right) \in \R^{\numdelays \cdot \latentsize}$.

\textbf{Temporal Chunking.} KALIKO temporally chunks the full-resolution data into groups of $n_x$ states, each corresponding to a sub-latent vector as follows:
\eqs{
    \latentdelay_t \leftrightarrow (\statechunk_{\tau}, \dots, \statechunk_{\tau+\chunksize-1}),
}
where $\tau$ is the original time index satisfying $t = \lfloor\tau / \chunksize\rfloor$ and $\statechunk_\tau$ is a raw data point. Thus, $\decoder$ decodes $\numdelays$ sub-latent vectors into a contiguous sequence of $\numdelays \cdot \chunksize$ raw data points:
\eqs{
    \decoder(\underbrace{\latentdelay_{t:(t+\numdelays-1)}}_{\latent_t}) = \underbrace{\statechunk_{\tau:(\tau + \chunksize \cdot \numdelays - 1)}}_{\state_t}.
}
Temporal chunking greatly speeds up training, as it reduces the number of steps in each training stage by a factor of $\chunksize$, until the cost of memory and linear system solves associated with a large data dimension $n$ outweighs the speedup. 

\textbf{Decoder Design.} The decoder $\decoder_\psi$ is a convolutional neural network that interleaves depthwise convolutions across the delay dimension (mixing $\latentdelay$'s) with GELU-activated MLPs, which allows $\decoder$ to mix sub-latent vectors across many time steps to decode long sequences of raw observations.
We found two such blocks sufficient for all experiments.

\begin{figure*}
    \vspace{0.15cm}
    \begin{center}
        \includegraphics[width=\linewidth]{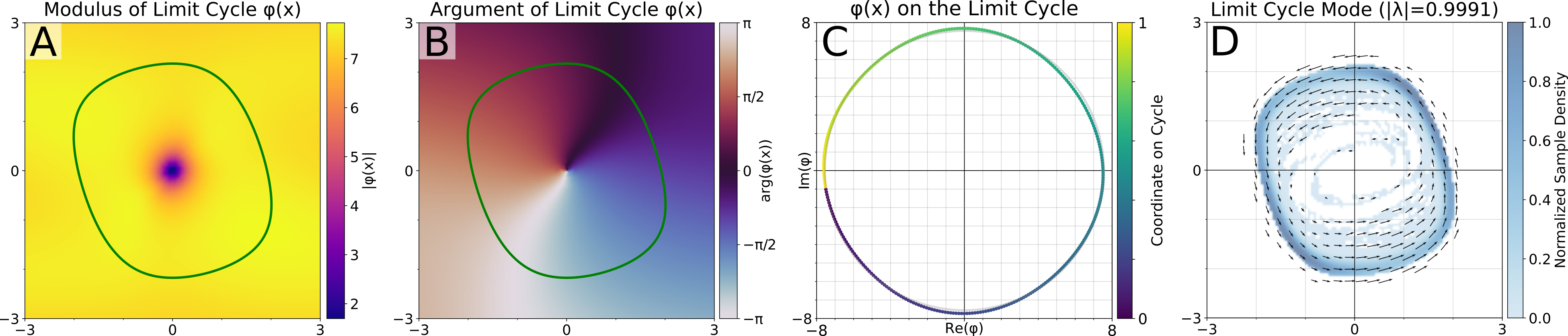}
    \end{center}
    \caption{\textbf{Limit Cycle Eigenfunction for the VDP System.} \textbf{(A/B)} Without parameterizing an encoder, KALIKO implicitly recovers an eigenfunction associated with the Van der Pol system's limit cycle (overlaid in green) with $|\lambda| \approx 1$. \textbf{(C)} Evaluating $\varphi(x)$ along the limit cycle traces out a nearly perfect circle in the complex plane, showing that KALIKO entirely captures the limit cycle dynamics into this eigenfunction. \textbf{(D)} The Koopman mode associated with this eigenvalue corresponds to a vector field that drives trajectories onto the cycle.}
    \label{fig:vdp_eigen}
    \vspace{-0.5cm}
\end{figure*}


\section{Analyzing KALIKO Embeddings}\label{sec:embeddings}

\subsection{Reconstruction Performance}

As discussed in Sec. \ref{sec:prelims_kalman}, the Kalman filter/smoother can be viewed as an encoder on sequences of data $\state_{1:T}$. We study KALIKO's autoencoding capabilities on four systems for visualization purposes: the Van der Pol oscillator, pendulum, Duffing oscillator, and a two-limit cycle Hopf-Bautin system.

\newcommand{\syscell}[2]{%
  \refstepcounter{equation}\label{#1}%
  \begingroup
    \everymath{\displaystyle}
    $\left\{
      \begin{aligned}
        #2
      \end{aligned}
    \right.$%
    \hfill(\theequation)%
  \endgroup
}

\begin{table}[h]
  \centering
  \setlength{\tabcolsep}{6pt}
  \renewcommand{\arraystretch}{1.25} 
  \footnotesize
  \begin{tabular}{|>{\raggedleft\arraybackslash}m{0.25\columnwidth}|m{0.62\columnwidth}|}
    \hline
    Van der Pol &
      \syscell{eq:vdp}{\label{eq:vdp}
        \dot{x}_1 &= x_1 - x_1^3/3 - x_2 \\
        \dot{x}_2 &= x_1
      } \\[6pt]
    \hline
    Pendulum &
      \syscell{eq:pendulum}{\label{eq:pendulum}
        \dot{x}_1 &= x_2 \\
        \dot{x}_2 &= \sin(x_1)
      } \\[6pt]
    \hline
    Duffing &
      \syscell{eq:duffing}{\label{eq:duffing}
        \dot{x}_1 &= x_2 \\
        \dot{x}_2 &= x_1 - \delta\,x_2 - x_1^3
      } \\[6pt]
    \hline
    Hopf-Bautin (Polar) &
      \syscell{eq:hopf_bautin}{\label{eq:hopf_bautin}
        \dot{r}     &= r e^{(1-2r^2)/2}\,(r^2 - r^4 - 3/16) \\
        \dot{\theta}&= 1
      } \\
    \hline
  \end{tabular}
  \label{tab:toy_systems}
  \caption{
    Systems used for embedding analysis. We learn discrete-time dynamics despite presenting these systems in continuous time for clarity.
  }
\end{table}

We plot reconstructed trajectories on a heatmap of reconstruction errors (Fig. \ref{fig:toy_recon}). 
KALIKO accurately reconstructs the original data (MAE $<10^{-2}$), verifying that the Kalman filter and smooth function as an effective encoder $\encoder$. Moreover, all models used identical hyperparameters, demonstrating KALIKO's robustness to features of nonlinear systems like fixed points, limit cycles, and multiple attractors.

\subsection{Analyzing KALIKO's Eigenfunctions}

Sec. \ref{sec:prelims_koopman} introduced a key property, $\koop$-invariance, in \eqref{eq:koop_closure}. Observe that the \textit{eigenfunctions} $\varphi$ of $\koop$ satisfy \eqref{eq:koop_closure}, since
\eqs{\label{eq:koop_eigenfunction}
    \lambda\varphi(\state_t) = \parens{\koop\varphi}(\state_t),\quad \lambda \in \mathbb{C},
}
thus giving coordinates on which $\koop$ acts linearly, making them central objects of study in the literature \cite{brunton2022_koopmanreview}. A left eigenpair $(\lambda, \mathbf{w})$ of the approximation $\latentdyn$ of $\koop$ (and $\encoder$) implies an eigenfunction $\varphi(\state) := \mathbf{w}^\top \encoder(\state)$, because
\eqsnn{
    \parens{\koop \varphi}(\state_t) &= \varphi(\statedyn(\state_t)) = \varphi(\state_{t+1}) = \mathbf{w}^\top \encoder(\state_{t+1}) \\
                       &= \mathbf{w}^\top \latentdyn \encoder(\state_t) = \lambda \mathbf{w}^\top\encoder(\state_t) = \lambda \varphi(\state_t).
}
Thus, we examine the eigenfunctions of $\latentdyn_\theta$ via eigendecomposition, allowing us to interpret the dynamics captured by KALIKO's implicit encoder, the Kalman filter and smoother. We consider two case studies.

\textbf{Van der Pol.} The Van der Pol (VDP) system \eqref{eq:vdp} has a globally-stable limit cycle (Fig. \ref{fig:toy_recon}, top left). 
We identify a complex eigenpair $(\lambda, \mathbf{w})$ of $\latentdyn_\theta$ with $|\lambda| \approx 1$, implying a steady-state oscillatory latent mode. 
We approximate the eigenfunction $\varphi(\state)$ by evaluating $\mathbf{w}^\top \encoder(\state)$ over latent beliefs in the plane, plotting VDP's limit cycle (green) against the modulus and argument of $\varphi(\state)$ in Fig. \ref{fig:vdp_eigen}A/B.

We can now check whether the limit cycle is described by $\varphi(\state)$. First, we parameterize the (unit-modulus) eigenvalue as $\lambda=e^{i\omega}$ and let $\latent_0$ be an arbitrary fixed latent element of the corresponding eigenspace. Observe that the corresponding latent trajectory must have constant modulus:
\eqsnn{
    |\latent_t| = |\lambda^t| \cdot |\latent_0| = |e^{it\omega}| \cdot |\latent_0| = |\latent_0|.
}
Moreover, the argument along the latent trajectory must evolve linearly (mod $2\pi$):
\eqsnn{
    \arg(\latent_t) = \arg(\lambda^t \latent_0) = t\omega + \arg(\latent_0) \quad (\text{mod } 2\pi).
}
Thus, if $\varphi(\state)$ evaluated on the limit cycle traces out a single revolution of a circle in the complex plane, we can conclude that KALIKO \textit{encodes the limit cycle dynamics entirely in this mode}, which is confirmed in Fig. \ref{fig:toy_recon}C.

We also examine the time-invariant directions in $\mathcal{X}$ given by the associated \textit{right eigenvector} of $\latentdyn_\theta$, or the \textit{Koopman mode} \cite{brunton2022_koopmanreview}. To do so, we encode a set of trajectories, project them onto the limit cycle eigenspace, and decode them, resulting in Fig. \ref{fig:vdp_eigen}D. The Koopman mode shows the attraction of the limit cycle in a neighborhood around it, verifying that KALIKO has learned VDP's dominant dynamical behavior.

\begin{figure*}
    \vspace{0.15cm}
    \begin{center}
        \includegraphics[width=\linewidth]{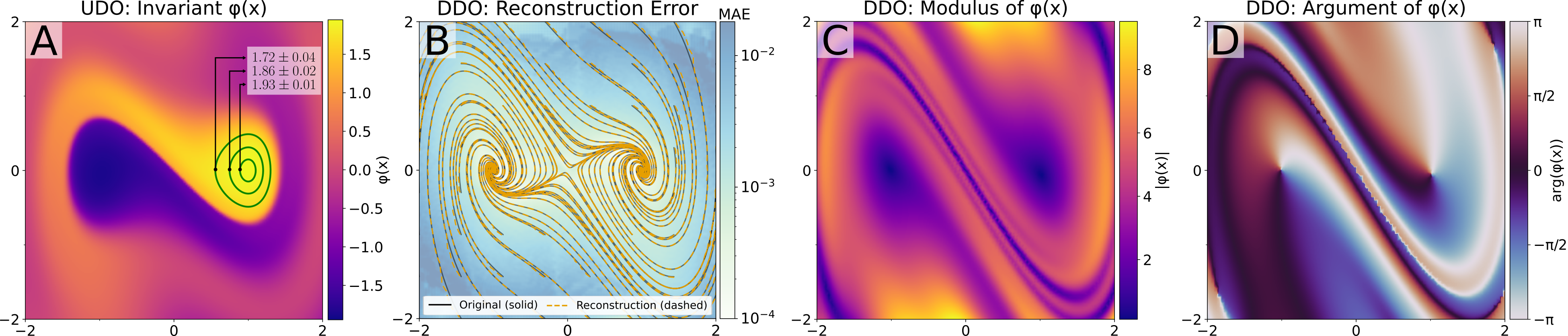}
    \end{center}
    \caption{\textbf{Eigenfunctions for the Undamped and Damped Duffing Oscillator.} \textbf{(A)} KALIKO recovers an eigenfunction $\varphi(\state)$ with $\lambda \approx +1$ capturing the invariance of systems with continuous spectra like the \textit{undamped Duffing oscillator} (UDO). The value of $\varphi(x)$ is nearly constant along each of three distinct cycles (green) with little variation (shown are mean and stdev). Across cycles, the value clearly changes, corresponding to energy invariance. \textbf{(B)} Conversely, KALIKO also reconstructs the \textit{damped Duffing oscillator} (DDO) with no adjustments. \textbf{(C/D)} KALIKO recovers an eigenfunction capturing the attractive ``spiral'' behavior about the wells at $(\pm1,0)$, reproducing the results from explicit methods in prior work \cite[Fig. 5]{otto2019_linearlyrecurrentautoencodernetworkslearning}.}
    \label{fig:duffing_eigen}
    \vspace{-0.5cm}
\end{figure*}

\textbf{Duffing Oscillators.} The undamped Duffing oscillator (UDO, $\delta=0$ in \eqref{eq:duffing}) has a saddle at the origin and two neutrally stable wells at $(\pm1,0)$. Because the oscillation frequency depends on energy (i.e., continuous spectrum \cite{lusch2018_deepkoopman}), no finite embedding space can recover a global phase eigenfunction; instead, invariants (e.g., energy) correspond to $\lambda\approx 1$. Consistent with this, KALIKO recovers an eigenfunction $\varphi$ that is nearly constant along each closed orbit and varies across energy levels (Fig.~\ref{fig:duffing_eigen}A), capturing UDO's energy invariance. Adding damping (DDO, $\delta=1$) eliminates the continuous spectrum, leaving the saddle at the origin and making $(\pm1,0)$ asymptotically stable. Here, KALIKO identifies a contractive mode with $|\lambda|<1$ whose eigenfunction $\varphi$ tends to $0$ at the three equilibria and whose phase clearly depicts the attractive spirals (Fig.~\ref{fig:duffing_eigen}B–D), in line with Koopman analyses in prior work \cite[Fig. 5]{otto2019_linearlyrecurrentautoencodernetworkslearning}.

\textbf{Takeaway.} Despite using an \textit{implicit} encoder, KALIKO reproduces the expected Koopman structure on these canonical systems (unit-modulus phase on the VDP limit cycle, energy-like invariants for UDO, and contractive modes near DDO equilibria) using \textit{identical hyperparameters} with no system-specific tuning. These results illustrate that KALIKO's implicit encoding does not preclude typical Koopman analysis; rather, it facilitates it by stably recovering expressive, interpretable representations without sensitive handcrafted features or model tuning.

\section{Experiments}\label{sec:experiments}

\subsection{Simulated Ocean Data}\label{sec:ocean_data}
We now study the predictive capacity of KALIKO and its application to a closed-loop stabilization control problem. To do so, we simulate the motion of a barge in various ocean conditions, yielding thousands of trajectories in $SE(3)$. The data are generated using \verb|capytaine| \cite{ancellin2019_capytaine}, which solves a linear-flow potential PDE in the frequency domain using the boundary element method. The irregular sea state is generated from the JONSWAP spectrum \cite{hasselmann1973_JONSWAP} by randomly sampling the significant wave height, peak period, and peak enhancement. For a set of discrete frequencies, we solve radiation and diffraction problems to assemble the hydrodynamic load on the moving body, which we then convert to a time series of motion by inverse Fourier transform. All parameter ranges were chosen such that the computed solutions were numerically stable, yielding realistic solutions. For more detail on hydrodynamics, we refer the reader to \cite{newman1977_marine_hydrodynamics}.

\begin{figure*}[th]
    \vspace{0.15cm}
    \begin{center}
        \includegraphics[width=\linewidth]{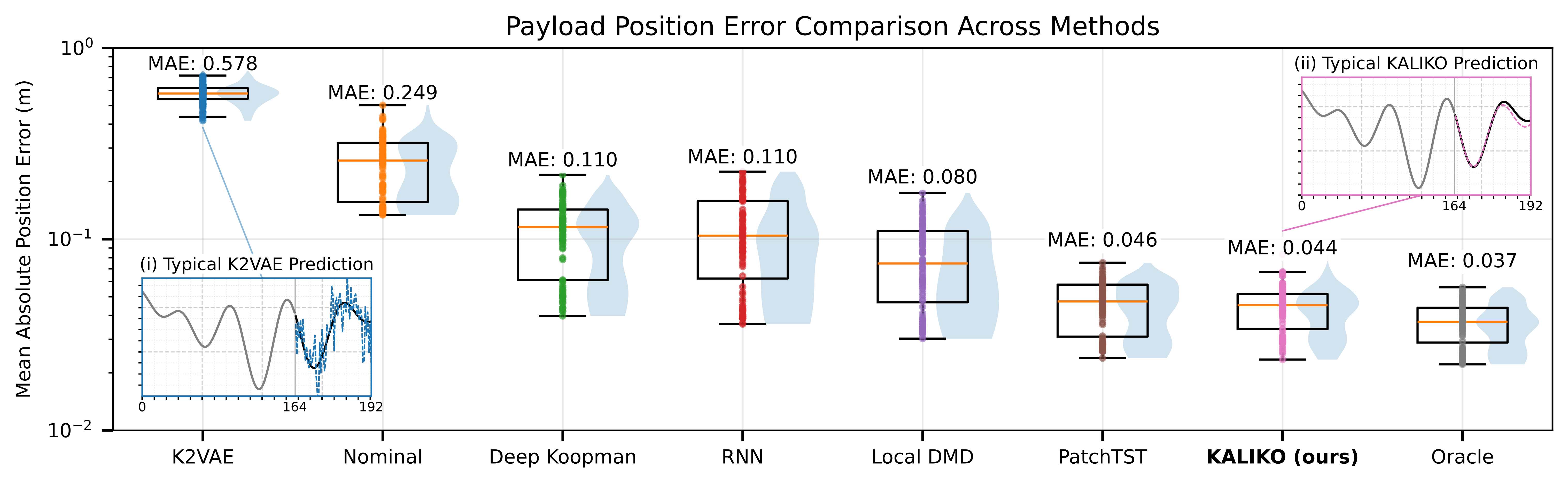}
    \end{center}
    \caption{\textbf{Control Results.} KALIKO outperforms all baselines on the control task with the least variance and near-oracle performance. Insets show one dimension of a hold-out trajectory. Gray denotes input context, black denotes ground truth prediction data. (i) K2VAE's poor performance can be explained by its overly-noisy predictions (blue, dashed). (ii) KALIKO's smooth prediction trajectories enable much more stable closed-loop control (pink, dashed).}
    \label{fig:sim_results}
    \vspace{-0.5cm}
\end{figure*}

\subsection{Baselines and Methodological Details}\label{sec:baselines}

We compare KALIKO to both Koopman-specific and general time series prediction baselines.

\textbf{Koopman Baselines.} We have 3 Koopman baselines. The \textit{Deep Koopman net} \cite{lusch2018_deepkoopman} is an autoencoding-based method that has the ability to parameterize continuous eigenvalue spectra. It is trained for long-horizon predictions with a multistep prediction consistency loss along with autoencoding losses, giving a canonical baseline for Koopman autoencoders. The \textit{Koopman-Kalman Variational Autoencoder} (K2VAE) \cite{wu2025_k2vae} claims state-of-the-art performance in probabilistic time series prediction. K2VAE predicts a ``residual'' global Koopman operator $\latentdyn_\text{glo}$ added to a locally-fitted operator derived from DMD, $\latentdyn_\text{loc}$, yielding the model $\latentdyn = \latentdyn_\text{loc} + \latentdyn_\text{glo}$. It uses an auxiliary Kalman filter to estimate and drive errors to 0.
Since K2VAE assumes that $\latentdyn_\text{loc}$ is a good prior about which to learn a residual, we implement a simple baseline where we fit a \textit{local DMD} model at each time step on the context $T_\text{in}$ using delay embeddings as the Koopman embeddings. Several works have explored variations of this idea in an MPC-style context for modeling effects like fluid flows \cite{bao2019_dmd_local_fluids, li2025_adaptivelocaldmd}. We used the open-source implementations and settings of the baselines for fair comparison.

\textbf{Time Series Baselines.} We have 2 general time series baselines. \textit{Recurrent neural networks} (RNNs) \cite{sherstinsky2020_rnns} are classic autoregressive models well-suited for dynamical prediction. RNNs can be viewed as ``generalized'' filters, making them a natural baseline: they ingest data, update an internal state, and can generate prediction sequences. \textit{PatchTST} \cite{nie2023_patchtst} is a transformer that consumes data as temporal ``patches'', inspired by vision transformers \cite{dosovitskiy2021_vit}. PatchTST is a popular method for general time series modeling, as it requires little domain-specific knowledge to yield reasonable predictions.

\textbf{Data Preprocessing.} For all methods, we map all poses in $SE(3)$ to $\mathfrak{se}(3)$, the tangent space at the identity, using the $\text{Log}$ map. This makes the data Euclidean and enables geometrically consistent pose predictions. Second, we normalize the data with a running estimate of the mean and standard deviation along each axis, which is updated during training. At inference time, we fix these statistics to prevent leakage.

\subsection{Prediction Performance}\label{sec:experiments_prediction}

In \textit{open-loop prediction}, each model is supplied an input context of $T_\text{in}=128$ data points and predicts $T_\text{out}=64$ steps. If applicable, methods are trained for 164k steps with a batch size of 32, and we report prediction errors in $\mathfrak{se}(3)$ averaged over all times, dimensions, and batches.

\begin{table}[h]
    \centering
    \begin{tabular}{|c||c|c|}
    \hline
    Method & MSE ($\downarrow$) & MAE ($\downarrow$) \\
    \hline
    Deep Koopman & 0.0019 & 0.025 \\
    K2VAE & 0.0015 & 0.026 \\
    Local DMD & 0.0005 & 0.010 \\
    RNN & 0.0003 & 0.010 \\
    PatchTST & 0.0004 & 0.011 \\
    \hline
    \textbf{KALIKO (ours)} & \textbf{0.0002} & \textbf{0.008} \\
    \hline
    \end{tabular}
    \caption{Performance on open-loop wave prediction.}
    \label{tab:wave_prediction_results}
\end{table}

In Table \ref{tab:wave_prediction_results}, KALIKO outperforms all methods using a global linear latent prediction model. In contrast, the RNN and PatchTST strongly leverage nonlinear models, and K2VAE and local DMD rely heavily on locally-fitted models rather than a global linear operator. Surprisingly, local DMD outperforms both Deep Koopman and K2VAE, which we attribute to the sensitivity of autoencoder-based Koopman methods to their learned embeddings, as noted in Sec. \ref{sec:related_work}, and K2VAE's noisy predictions (see Fig. \ref{fig:sim_results}).

\subsection{Ablations}\label{sec:ablations}

\textbf{Delay Length.} KALIKO uses a delay length $n_z=4$ for its latent dynamics. We perform two ablations: letting $n_z=1$ (no delay) and increasing to $n_z=6$. We find that the delay is crucial for performance, as the MAE nearly doubles without it. Conversely, using a higher delay does not improve performance, justifying the choice $n_z=4$.

\textbf{Decoder Architecture.} We replaced KALIKO's convolutional decoder with a simple MLP that performs no depthwise mixing, instead mapping each sub-latent vector $\latentdelay_t$ to a distinct temporal chunk. We find this only slightly harms performance, showing KALIKO's robustness to architectural design decisions.

\textbf{Learnable Priors.} We evaluate KALIKO's performance when fixing the prior to $(0,I)$. This substantially degrades performance, which we hypothesize is because the scale and anisotropy of the prior are key for allowing KALIKO to learn the most expressive possible embedding space geometry.

\begin{table}[h]
    \centering
    \begin{tabular}{|c||c|c|}
    \hline
    Ablation & MSE ($\downarrow$) & MAE ($\downarrow$) \\
    \hline
    $n_z=4 \rightarrow n_z=1$ & 0.0007 & 0.015 \\
    $n_z=4 \rightarrow n_z=6$ & 0.0002 & 0.008 \\
    Conv $\rightarrow$ MLP Decoder & 0.0003 & 0.009 \\
    Learnable $\rightarrow$ Fixed Latent Prior & 0.0005 & 0.012 \\
    \hline
    \textbf{KALIKO (Original)} & \textbf{0.0002} & \textbf{0.008} \\
    \hline
    \end{tabular}
    \caption{Ablations: effect on KALIKO's prediction performance.}
    \label{tab:ablations}
    \vspace{-0.5cm}
\end{table}

\subsection{Control Performance}\label{sec:experiments_control}
\begin{figure}[t]
    \begin{center}
        \includegraphics[width=\linewidth]{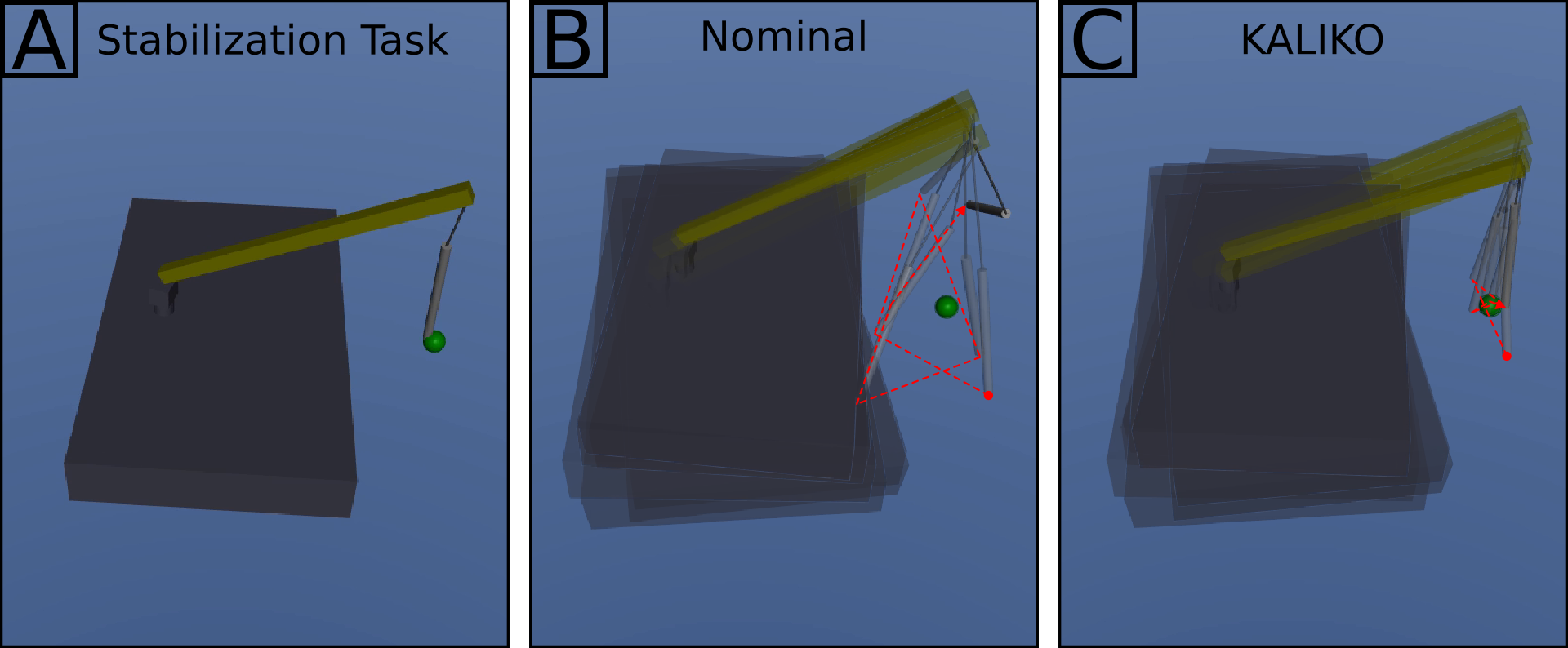}
    \end{center}
    \caption{\textbf{(A)} We aim to stabilize the tip of the payload to a goal location (green) by compensating for motion induced by realistic waves. 
    A crane controls the payload by rotating its base and boom and/or reeling its winch.
    \textbf{(B)} The nominal controller poorly regulates the payload, as seen in this 3 second snapshot (red dashed lines). \textbf{(C)} KALIKO stabilizes the payload at near-oracle levels in the presence of the same waves.}
    \label{fig:control_sim}
    \vspace{-0.4cm}
\end{figure}

We evaluate all methods on a simulated control task: stabilizing a cable-suspended payload with a barge-mounted crane (see Fig. \ref{fig:control_sim}).
The task is challenging because the barge and crane are severely underactuated: the free-body motion is driven by the ocean waves, and the crane's payload height is controllable only via a damped winch.
We need not model any control dynamics, since the crane is controlled with sampling-based MPC, which selects controls by choosing high-performance candidates under forward simulations of the system. Thus, the predictive models in Sec. \ref{sec:experiments_prediction} suffice.

We evaluate all methods on 100 random
wave simulations, feeding their predictions to an MPPI controller implemented using the open-source toolbox \verb|judo| \cite{li2025_judo}.
We calibrate performance against two other methods: an \textbf{oracle} supplied with ground-truth future barge motion, and a \textbf{nominal} stabilizing controller that ignores the wave disturbances.
Since the controller is stochastic, the oracle allows us to measure the error induced by its sampling noise. We report mean absolute error in the payload target position (see Fig. \ref{fig:sim_results}).

KALIKO outperforms all methods with near-oracle performance and the tightest variance, showing that its superior open-loop prediction translates to high-performance closed-loop control.
However, prediction performance is not a perfect indicator of control performance, as evidenced by local DMD outperforming the RNN.
We attribute this gap to the role of state estimation, which is as crucial as prediction in control.
Trained explicitly for filtering, KALIKO maintains an accurate latent belief in closed loop, whereas autoencoding-based Koopman methods and the RNN are trained only for prediction, and their internal states drift.
Conversely, local DMD and PatchTST, which are not latent-variable methods, perform well because they operate directly on (patches of) raw data; with reasonably accurate short-horizon predictions and fast enough feedback, the controller can effectively stabilize the system.


\section{Conclusion}\label{sec:conclusion}

We introduced Kalman-Implicit Koopman Operator (KALIKO) learning, which leverages Kalman filtering to implicitly compute Koopman embeddings and linear latent dynamics without explicitly parameterizing an encoder. KALIKO learns highly interpretable representations, achieves superior open-loop prediction, and outperforms all baselines on a challenging closed-loop control task, with near-oracle performance. In sum, KALIKO turns Bayesian filtering into a practical Koopman encoder, yielding performant embeddings that require no handcrafting or expert domain knowledge.

\bibliographystyle{unsrt}
\bibliography{references}

\end{document}